\documentclass{article}

 \usepackage[dblblindworkshop, final]{neurips_2026}
\workshoptitle{Geometric Distributional Deep Learning (GDDL)}

\usepackage[utf8]{inputenc} 
\usepackage[T1]{fontenc}    
\usepackage{hyperref}       
\usepackage{url}            
\usepackage{booktabs}       
\usepackage{amsfonts}       
\usepackage{nicefrac}       
\usepackage{microtype}      
\usepackage{xcolor}         
\usepackage{enumitem}
\usepackage{bbm}
\usepackage[table]{xcolor}
\usepackage{arydshln}

\usepackage{lmodern}  
\usepackage{amsmath,amssymb,amsthm}
\usepackage{graphicx}
\usepackage{hyperref}
\usepackage{xcolor}\usepackage{natbib}
\usepackage{tikz}
\usetikzlibrary{arrows.meta,shapes.geometric}

\newcommand{\model}{\textsc{Hermes}}

\newcommand{\hmicro}{h_{\mathrm{micro}}}
\newcommand{\hmacro}{h_{\mathrm{macro}}}
\newcommand{\hout}{h_{\mathrm{out}}}

\newcommand{\Kf}{K_f}
\newcommand{\Knf}{K_{nf}}

\title{%
  \textbf{Dual-Scale Relational Graph Transformers \\ for Ecosystem-Aware Fraud Detection}
}

\author{
\small
Mohsen Nayebi Kerdabadi$^{1}$ \hspace{0.5em}
Xinrou Li$^{2}$ \hspace{0.5em}
Yao Xiao$^{2}$ \hspace{0.5em}
Zijun Yao$^{1}$ \hspace{0.5em}
Xin Sun$^{2}$
\\[0.5em]
$^{1}$University of Kansas, Lawrence, KS, USA
\\
$^{2}$AI Foundations, Capital One, McLean, VA, USA
\\[0.3em]
\texttt{\{xinrou.li,yao.xiao,xin.sun\}@capitalone.com}
\\
\texttt{\{mohsen.nayebi,zyao\}@ku.edu}
}

\begin{document}

\maketitle

\begin{abstract}
Account takeover (ATO) fraud is a growing threat to digital banking, requiring effective detection while minimizing friction for legitimate customers. Production systems predominantly rely on tabular models that score sessions in isolation, discarding the relational structure of the underlying interaction network. Although graph-based models exploit relationships among sessions and network entities, they primarily reason over local neighborhoods and therefore capture only part of the problem: fraud risk depends jointly on the local relational structure surrounding a session and the evolving global state of the fraud ecosystem.
We present \model{} (\textbf{HE}terogeneous \textbf{R}elational \textbf{M}icro--macro graph transformer \textbf{E}ncoder for high-risk \textbf{S}essions), a dual-scale architecture that jointly models these complementary scales of information. \emph{Micro-GT} captures local heterogeneous graph structure through structured, relation-aware attention over a temporally safe session neighborhood. Complementing this local representation, \emph{Macro-GT} models ecosystem-level context using non-anticipative climate tokens that summarize fraud dynamics, platform shifts, and infrastructure reuse, together with adaptive class prototypes that track representative fraud and benign session patterns over time.
Evaluated on more than 130 million high-risk transaction sessions from a leading U.S. financial institution, \model{} consistently outperforms production and strong graph-based baselines, achieving a 44.44\% relative reduction in customer friction and a 24.66\% relative improvement in fraud recall over the production system. Ablation and temporal-stability analyses further demonstrate complementary gains from local relational modeling and global ecosystem context across changing fraud regimes.

\end{abstract}

\section{Introduction}
\label{sec:intro}

Account takeover (ATO) fraud is a growing threat to digital financial services, causing substantial financial losses and eroding customer trust. In an ATO attack, an adversary gains access to a legitimate user's account to conduct unauthorized activity or obtain sensitive information. Effective detection therefore presents a fundamental operational tradeoff: increasing fraud recovery often requires challenging more legitimate customers, introducing friction that degrades user experience \citep{doerfler2019evaluating}. As attack strategies become increasingly sophisticated, practical fraud detection requires models that improve fraud recovery while minimizing unnecessary customer friction.

ATO fraud is inherently a structured and non-stationary prediction problem. Sessions are connected through a heterogeneous interaction graph of shared accounts, devices, and network identifiers, creating relation-specific and temporal dependencies that session-level features alone cannot capture \citep{kerdabadi2025spatio}.  At the same time, the risk associated with a local graph pattern depends on the broader fraud ecosystem, which evolves as attack volume, platform concentration, and infrastructure reuse shift over time. Reliable detection therefore requires reasoning jointly over the local relational structure surrounding a session and the evolving global ecosystem state in which that structure occurs. This challenge is further compounded by extreme class imbalance and delayed fraud adjudication, requiring models to operate non-anticipatively using only information available at scoring time.

Existing fraud-detection models capture these complementary sources of information only partially. Production tabular systems score sessions largely in isolation, discarding relational dependencies among connected entities. Graph neural networks \citep{hamilton2017inductive, 10.14778/3494124.3494128, xiang2023semi} improve upon this by aggregating information from local neighborhoods, and temporal label propagation \citep{kerdabadi2025spatio} can further incorporate confirmed fraud signals from connected sessions. However, conventional message passing can struggle to preserve informative dependencies across multiple hops because repeated aggregation may induce oversmoothing and oversquashing \citep{ICLR2026_fa1ef433}. Generic aggregation can also blur the distinct semantics of account, device, and network relations and inadequately capture higher-order interactions across heterogeneous paths within a multi-hop neighborhood. Moreover, models restricted to local neighborhoods remain largely unaware of ecosystem-level regime changes. Consequently, existing approaches lack a unified mechanism for combining expressive local relational modeling with the evolving global context in which that structure should be interpreted.

We present \model{}: \textbf{HE}terogeneous \textbf{R}elational \textbf{M}icro--macro Graph Transformer \textbf{E}ncoder for high-risk transaction \textbf{S}essions, a dual-scale architecture that addresses these limitations by jointly modeling local relational structure and global ecosystem context. \emph{Micro-GT} applies a relational graph transformer over a temporally safe sampled neighborhood, using four structured attention masks to preserve relation-specific semantics and capture higher-order interactions across heterogeneous paths. Complementing this local representation, \emph{Macro-GT} attends over non-anticipative ecosystem climate tokens that summarize evolving fraud dynamics, platform shifts, and infrastructure reuse, together with adaptive class prototypes that provide representative references for fraud and benign session patterns as they evolve. The two representations are fused and trained end-to-end for session-level fraud prediction.

Evaluated on more than 130 million high-risk transaction sessions collected over one year at a leading U.S. financial institution, \model{} consistently outperforms both the production gradient-boosted system and strong graph-based baselines. Relative to the production system, \model{} achieves a 44.44\% reduction in customer friction at fixed fraud recall and a 24.66\% improvement in fraud recall at a fixed flag rate. Ablation studies demonstrate complementary gains from the proposed components, while temporal evaluation across four held-out test months demonstrates robustness to distribution shift. Our main contributions are as follows:
\vspace{-0.4cm}
\begin{itemize}[leftmargin=*]
    \item We present \model{}, a dual-scale architecture that jointly reasons over local graph neighborhood structure and evolving global ecosystem-level distributional context, enabling session-level fraud prediction across two complementary scales within a unified framework.
    \vspace{-0.2cm}

    \item We propose Micro-GT, a local relational graph transformer that uses four structured attention masks to preserve heterogeneous relation semantics and capture higher-order interactions across temporally safe multi-hop neighborhoods.
    \vspace{-0.2cm}

    \item We propose Macro-GT, an ecosystem-scale transformer that attends over a rich set of
    non-anticipative daily climate tokens covering multiple fraud signal categories, augmented with adaptive class prototypes that provide representative references for fraud and benign session patterns.
    \vspace{-0.2cm}

    \item We conduct a production-scale evaluation on more than 130 million high-risk transaction sessions from a leading U.S. financial institution, with extensive ablations and temporal analyses demonstrating complementary gains from the local and global components and substantial improvements over the deployed production system.
\end{itemize}

\section{Problem Formulation}
\label{sec:problem}

\subsection{Heterogeneous Temporal Session Graph}

Let $\mathcal{S}$ denote the set of high-risk digital transaction sessions. Each session
$v \in \mathcal{S}$ is associated with a feature vector
$\mathbf{x}_v \in \mathbb{R}^{d}$, an occurrence timestamp $\tau(v)$, and a
binary fraud label $y_v \in \{0,1\}$. Because fraud labels are adjudicated with
delay, $y_v$ becomes observable only at $\tau(v) + \ell$, where $\ell > 0$
denotes the \emph{label lag}.

Let $\mathcal{R}$ denote the set of relation types corresponding to account,
device, and network identifiers, and let $\operatorname{id}_{r}(v)$ denote the
identifier of type $r \in \mathcal{R}$ associated with session $v$. We define
the relation-specific temporal predecessors of $v$ as
\begin{equation}
    \mathcal{N}_{r}^{-}(v)
    =
    \operatorname{TopK}_{\tau}
    \left(
    \left\{
        u \in \mathcal{S} :
        \operatorname{id}_{r}(u) = \operatorname{id}_{r}(v),\;
        \tau(v) - T \leq \tau(u) < \tau(v)
    \right\}
    \right),
    \label{eq:temporal_neighbors}
\end{equation}
where $T$ is the recency window and $\operatorname{TopK}_{\tau}(\cdot)$ retains
up to $K$ candidate sessions with the largest occurrence timestamps. 

For each destination session $v$, relation type $r$, and associated identifier
$a = \operatorname{id}_{r}(v)$, we instantiate a time-conditioned identifier
node $i_{r,a}^{\tau(v)}$, representing the state of identifier $a$ at time
$\tau(v)$. Rather than treating identifiers as static entities, this formulation
allows their representations to evolve as additional historical activity becomes
available. Each shared-identifier relation is represented as
\begin{equation}
    u \longrightarrow i_{r,a}^{(\tau(v))} \longrightarrow v,
    \qquad
    u \in \mathcal{N}_{r}^{-}(v),
    \quad
    a = \operatorname{id}_{r}(u) = \operatorname{id}_{r}(v).
    \label{eq:identifier_path}
\end{equation}
Thus, $i_{r,a}^{\tau(v)}$ represents the time-conditioned state of identifier
$a$, summarizing its history available when session $v$ is scored. For a target session $v_0$, we recursively apply
Eq.~\eqref{eq:temporal_neighbors} for $H$ session hops to construct the local
computation graph $\mathcal{G}(v_0)$. We define \emph{session-hop distance}
by treating each path
$u \rightarrow i_{r,a}^{(\tau(v))} \rightarrow v$ as one relational hop.
Accordingly, $v_0$ is assigned hop~0, its direct session predecessors hop~1,
and more distant predecessors increasing hop indices up to $H$. Because every
expansion satisfies $\tau(u) < \tau(v)$, $\mathcal{G}(v_0)$ is temporally
ordered and contains no future-session information.

\subsection{Temporally-Safe Node Representation}

Session nodes carry their session-level features $\mathbf{x}_v$. To incorporate
historical fraud information without label leakage, label-dependent features
are computed only from predecessors whose labels are observable when $v$ is
scored. For relation $r$, the lag-safe predecessor set is
\begin{equation}
    \mathcal{L}_{r}^{-}(v)
    =
    \left\{
        u \in \mathcal{N}_{r}^{-}(v) :
        \tau(u) + \ell \leq \tau(v)
    \right\}.
    \label{eq:lag_safe_neighbors}
\end{equation}
Thus, structural inclusion requires $\tau(u) < \tau(v)$, while label-dependent features additionally require $\tau(u) + \ell \leq \tau(v)$. Identifier-node features likewise use only information available at the destination timestamp. Full feature definitions are provided in Appendix~\ref{app:tokens}.

\subsection{Task Definition}

For a target session $v_0$, let $\mathbf{X}(v_0)$ denote the serve-time
features associated with its local computation graph, and let
$\mathbf{C}_{\tau(v_0)}$ denote the global ecosystem context available at
time $\tau(v_0)$. The session-level fraud detection task is
\begin{equation}
    \hat{y}_{v_0}
    =
    f_{\theta}\!\left(
        \mathcal{G}(v_0),
        \mathbf{X}(v_0),
        \mathbf{C}_{\tau(v_0)}
    \right)
    \in [0,1],
    \label{eq:fraud_prediction}
\end{equation}
where $\hat{y}_{v_0}$ estimates the probability that $v_0$ is fraudulent.
All local and global inputs are restricted to information available at
$\tau(v_0)$, ensuring consistent temporal constraints during training,
evaluation, and production inference.

\section{Method}
\label{sec:method}

\begin{figure}[t]
    \centering
    \includegraphics[width=0.9\linewidth]{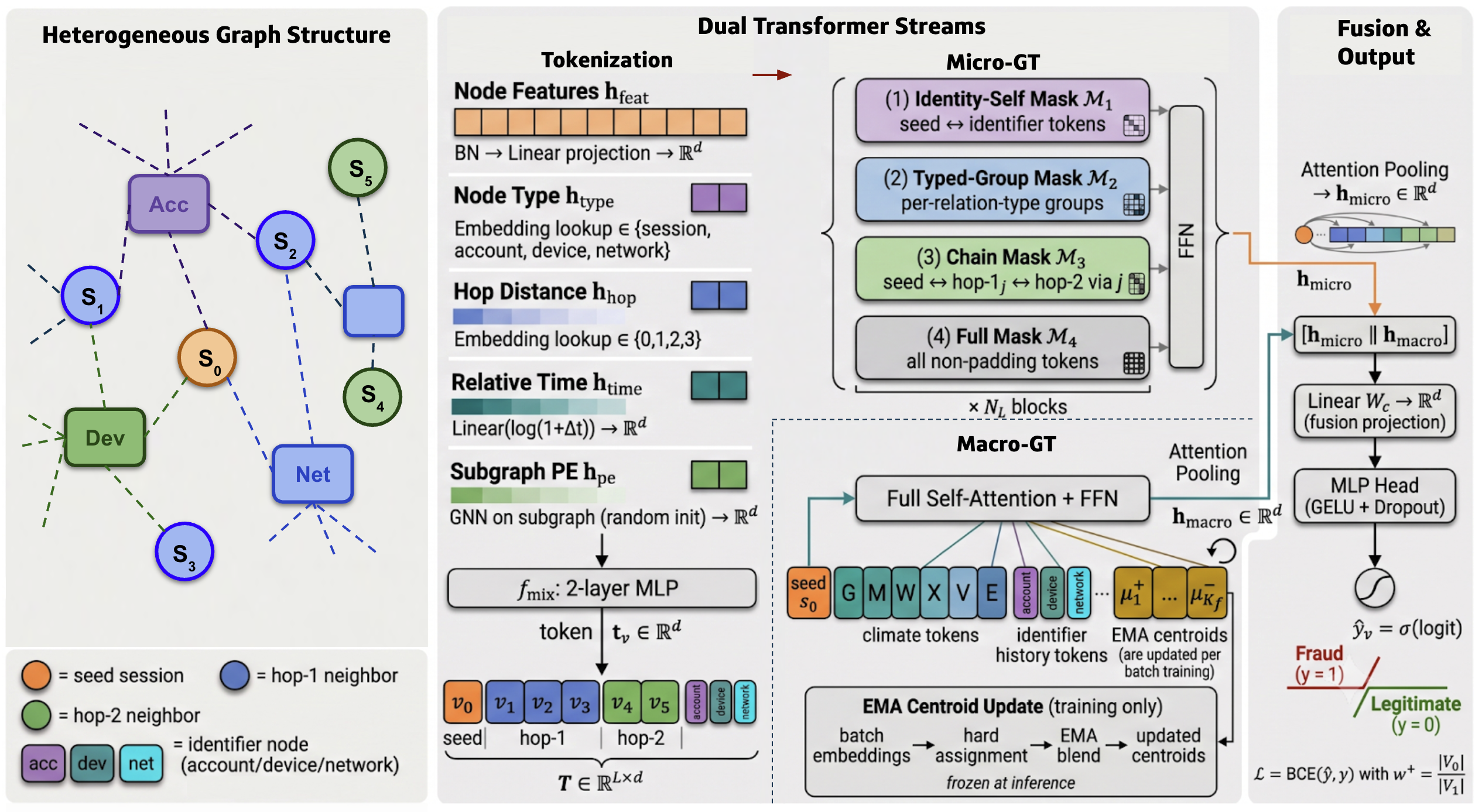}
    \caption{\model{} architecture. Micro-GT encodes the target session's
    local heterogeneous neighborhood through structured relational attention,
    while Macro-GT models non-anticipative ecosystem context together with
    adaptive fraud and benign reference prototypes. 
    }
    \label{fig:architecture}
    \vspace{-4mm}
\end{figure}

\model{} is a dual-scale architecture that represents each target session from
two complementary perspectives. \emph{Micro-GT} models the local relational
structure of the target-conditioned temporal graph, while \emph{Macro-GT}
captures the evolving ecosystem context available at the target's scoring
time. The two representations are fused through a learned projection and
optimized jointly for session-level fraud prediction.
Figure~\ref{fig:architecture} provides an overview.

\subsection{Local Graph Tokenization}
\label{sec:method:sampling}

For a target session $v_0$, Micro-GT operates on the target-conditioned
computation graph $\mathcal{G}(v_0)$ defined in
Section~\ref{sec:problem}. We instantiate $H=2$ session hops and retain fixed
budgets of at most $K_1$ hop-1 and $K_2$ hop-2 session predecessors, selected
according to the temporal sampling rule in Eq.~\eqref{eq:temporal_neighbors}.
Missing positions are padded and masked from all subsequent attention
operations.

Time-conditioned identifier states mediate relational paths throughout
$\mathcal{G}(v_0)$, but only those adjacent to the target are explicitly
tokenized. Specifically, for each relation $r\in\mathcal{R}$, we include
$i_{r,a_r}^{\tau(v_0)}$, where
$a_r=\operatorname{id}_{r}(v_0)$. Their serve-time feature construction is
detailed in Appendix~\ref{app:id_tokens}. Figure~\ref{fig:micro_tokenization}
illustrates the resulting local context.

Each tokenized session or identifier state $q$ is mapped to
$\mathbf{t}_q\in\mathbb{R}^{d}$ by combining feature, type, hop, temporal, and
structural encodings following~\citet{ICLR2026_fa1ef433}:
\begin{equation}
    \mathbf{t}_q
    =
    f_{\mathrm{mix}}\!\left(
        \mathbf{h}_{\mathrm{feat}}(q),\,
        \mathbf{h}_{\mathrm{type}}(q),\,
        \mathbf{h}_{\mathrm{hop}}(q),\,
        \mathbf{h}_{\mathrm{time}}(q,v_0),\,
        \mathbf{h}_{\mathrm{pe}}(q)
    \right),
    \label{eq:token_representation}
\end{equation}

where $f_{\mathrm{mix}}$ is a two-layer MLP projecting the concatenated
components to dimension $d$. The components are
\begin{equation} 
\begin{aligned} 
\mathbf{h}_{\mathrm{feat}}(q)
&=
\mathrm{MLP}_{\phi(q)}
\!\left(\mathrm{BN}_{\phi(q)}(\mathbf{x}_q)\right)
\in \mathbb{R}^{d}, \\
\mathbf{h}_{\mathrm{type}}(q) &= \mathbf{E}_{\mathrm{type}}[\phi(q)], \\ \mathbf{h}_{\mathrm{hop}}(q) &= \mathbf{E}_{\mathrm{hop}}[p(q,v_0)], \\ \mathbf{h}_{\mathrm{time}}(q,v_0) &= \mathbf{w}_t \log\!\left(1+\Delta\tau(q,v_0)\right)+b_t . 
\end{aligned} 
\label{eq:token_components} 
\end{equation}
Here, $\mathrm{MLP}_{\phi(q)}$ and $\mathrm{BN}_{\phi(q)}$ denote
type-specific feature projection and normalization, mapping raw node
features of potentially different dimensions to $d$; $\phi(q)$ denotes the node type and
$p(q,v_0)$ the session-hop index.
The target and its adjacent identifier states are assigned hop~0, while
hop-1 and hop-2 sessions receive indices 1 and 2, respectively. For session
tokens, $\Delta\tau(q,v_0)=\tau(v_0)-\tau(q)$; for target-adjacent identifier
states, $\Delta\tau(q,v_0)=0$. Finally,
$\mathbf{h}_{\mathrm{pe}}(q)$ is a permutation-equivariant structural
positional encoding produced by a lightweight GNN with randomized
initialization over $\mathcal{G}(v_0)$, following
\citet{ICLR2026_fa1ef433}.

Let $\{v_k^{(1)}\}_{k=1}^{K_1}$ and
$\{v_k^{(2)}\}_{k=1}^{K_2}$ denote the padded hop-1 and hop-2 session slots,
and let $a_j=\operatorname{id}_{r_j}(v_0)$. The Micro-GT input sequence is
\begin{equation}
\begin{aligned}
    \mathbf{Z}^{(0)}
    =
    \bigl[
        &\mathbf{t}_{v_0}
        \;\big|\;
        \mathbf{t}_{v_1^{(1)}},\ldots,\mathbf{t}_{v_{K_1}^{(1)}}
        \;\big|\;
        \mathbf{t}_{v_1^{(2)}},\ldots,\mathbf{t}_{v_{K_2}^{(2)}}
        \;\big|\; \\
        &\mathbf{t}_{i_{r_1,a_1}^{(\tau(v_0))}},\ldots,
        \mathbf{t}_{i_{r_{|\mathcal{R}|},a_{|\mathcal{R}|}}^{(\tau(v_0))}}
    \bigr]
    \in\mathbb{R}^{L\times d},
    \qquad
    L = 1 + K_1 + K_2 + |\mathcal{R}|.
\end{aligned}
\label{eq:micro_sequence}
\end{equation}
Padding positions remain masked throughout Micro-GT.

\begin{figure}[t]
    \centering
    \includegraphics[width=0.75\linewidth]{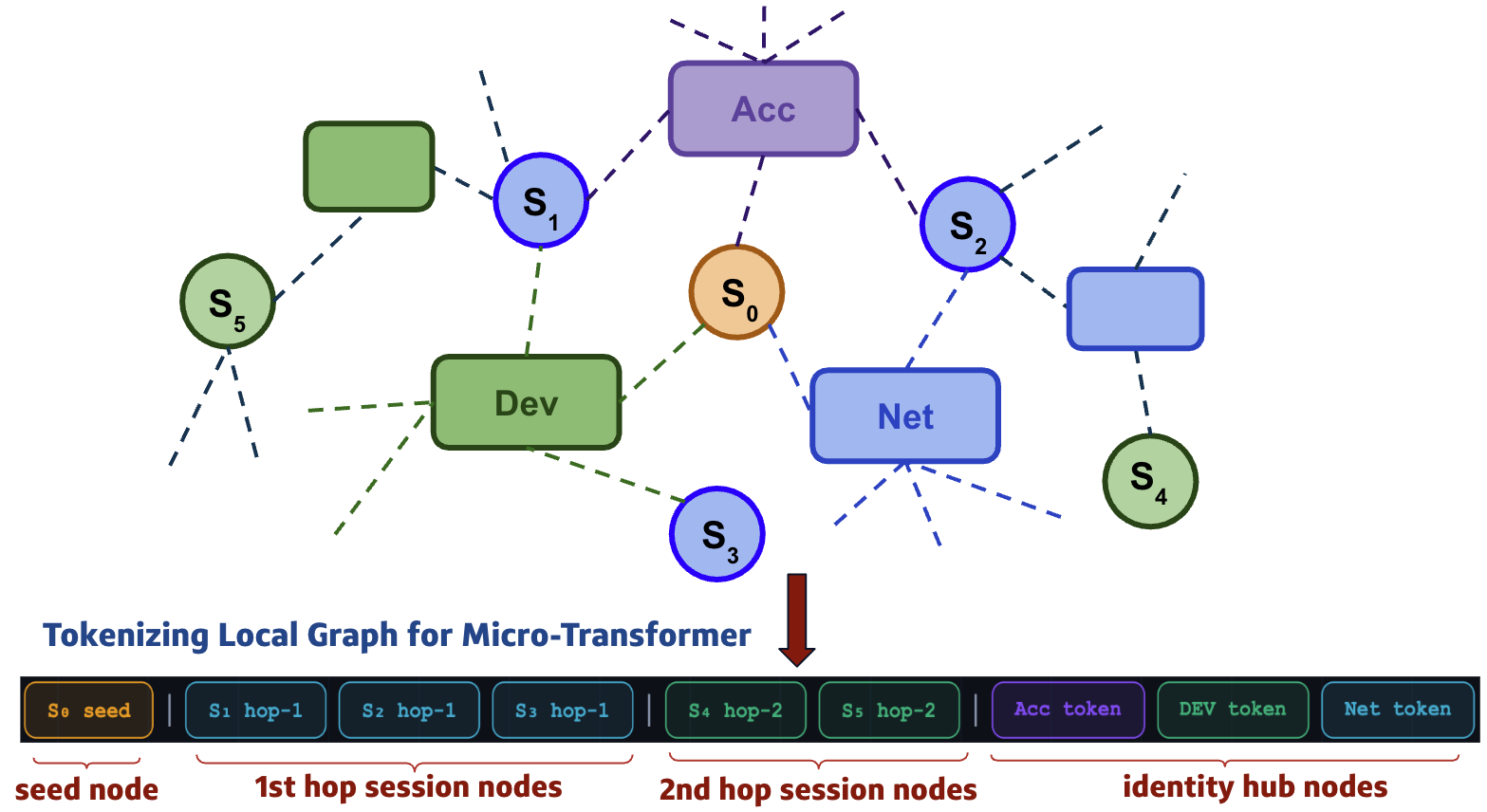}
    \caption{Micro-GT local graph tokenization with the target session, hop-1 and
hop-2 predecessors, and target-adjacent time-conditioned identifier states.}
    \label{fig:micro_tokenization}
    \vspace{-4mm}
\end{figure}

\subsection{Micro-GT: Local Relational Transformer}
\label{sec:method:micro}

\begin{figure}[t]
    \centering
    \includegraphics[width=\linewidth]{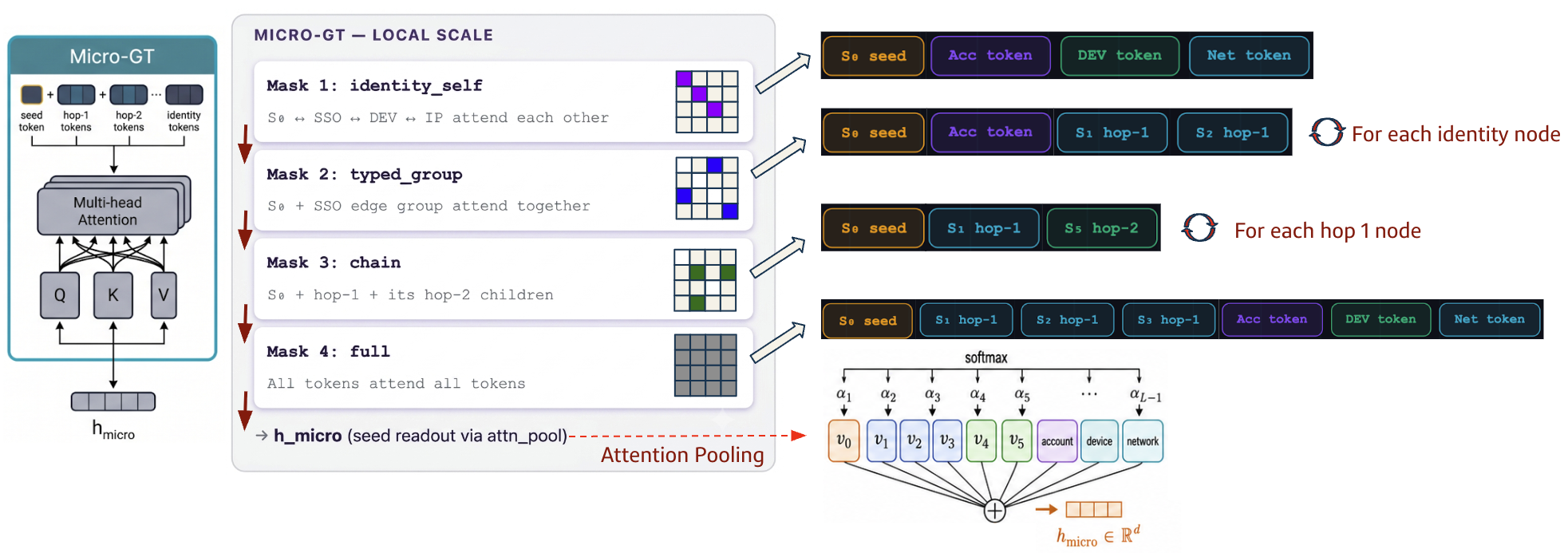}
    \caption{Micro-GT applies four sequential attention patterns over the local
    token sequence followed by seed-conditioned pooling
    to obtain $\mathbf{h}_{\mathrm{micro}}$.}
    \label{fig:micro_gt}
    \vspace{-4mm}
\end{figure}

Let $\mathbf{T}=\mathbf{Z}^{(0)}$ denote the token sequence from
Eq.~\eqref{eq:micro_sequence}. Rather than applying undifferentiated all-pair
attention, Micro-GT uses four structured attention masks applied sequentially,
each with independent query, key, and value projections. Each masked attention
pass follows the pre-norm residual form
\begin{equation}
    \mathbf{T}
    \leftarrow
    \mathbf{T}
    +
    \operatorname{Attn}_m
    \left(
        \operatorname{LN}(\mathbf{T}),
        M_m
    \right),
    \qquad
    \operatorname{Attn}_m(\mathbf{T},M_m)
    =
    \operatorname{softmax}
    \left(
        \frac{\mathbf{Q}_m\mathbf{K}_m^\top}{\sqrt{d_k}}
        + M_m
    \right)
    \mathbf{V}_m,
    \label{eq:micro_attention}
\end{equation}
where
$\mathbf{Q}_m=\mathbf{T}\mathbf{W}_m^Q$,
$\mathbf{K}_m=\mathbf{T}\mathbf{W}_m^K$, and
$\mathbf{V}_m=\mathbf{T}\mathbf{W}_m^V$.
Each $M_m\in\{0,-\infty\}^{L\times L}$ specifies the permitted interactions
for attention pass $m$. The four masks are illustrated in
Figure~\ref{fig:micro_gt}. 

\textbf{(1) Identity-Self mask $\mathbf{M}_1$.} The target token and its identifier tokens attend to one another, allowing the model to first establish the target's relational identity context. Let  $\mathcal{I}_0 = \left\{ v_0 \right\} \cup \left\{ i_{r,a_r}^{\tau(v_0)} : r\in\mathcal{R} \right\}$. Then $\mathbf{M}_1[i,j]=0$ iff both tokens $i$ and $j$ belong to $\mathcal{I}_0$.

\textbf{(2) Relation-group mask $\mathbf{M}_2$.} For each session token $q$, let $\rho(q)\subseteq\mathcal{R}$ denote the relation types it shares with the target. Two session tokens may interact when their relation sets overlap, while the target and identifier-context tokens remain accessible to all groups: \begin{equation} \mathbf{M}_2[i,j]=0 \iff \rho(i)\cap\rho(j)\neq\emptyset \;\; \text{or} \;\; i\in\mathcal{I}_0 \;\; \text{or} \;\; j\in\mathcal{I}_0 . \label{eq:relation_group_mask} \end{equation} This preserves relation-specific interactions while allowing sessions linked through multiple identifiers to participate in multiple groups. 

\textbf{(3) Chain mask $\mathbf{M}_3$.} For each hop-1 session, the target, that hop-1 session, and its associated hop-2 predecessors form an attention group. This explicitly models multi-hop relational chains while keeping paths through different hop-1 intermediaries separate. \paragraph{(4) Full-context mask $\mathbf{M}_4$.} All non-padding tokens attend to one another, enabling information exchange across relation groups and propagation chains after the structured passes.

The four passes are applied in order: identity-self, relation-group, chain, full-context, followed by a shared feed-forward sublayer with pre-norm residual. This constitutes one relational transformer block, and
\model{} stacks $N_L$ such blocks.

The final local representation
$\mathbf{h}_{\mathrm{micro}}\in\mathbb{R}^{d}$ is obtained through
seed-conditioned attention pooling:
\begin{equation}
    \mathbf{h}_{\mathrm{micro}}
    =
    \mathbf{t}_{v_0}^{(N_L)}
    +
    \sum_{i\neq v_0}
    \alpha_i\,\mathbf{t}_{i}^{(N_L)},
    \qquad
    \alpha_i
    =
    \frac{
        \exp\!\left(
            \mathbf{w}_r^\top
            [\mathbf{t}_{v_0}^{(N_L)} \| \mathbf{t}_{i}^{(N_L)}]
        \right)
    }{
        \sum_{j\neq v_0}
        \exp\!\left(
            \mathbf{w}_r^\top
            [\mathbf{t}_{v_0}^{(N_L)} \| \mathbf{t}_{j}^{(N_L)}]
        \right)
    },
    \label{eq:micro_pooling}
\end{equation}
where $\mathbf{w}_r\in\mathbb{R}^{2d}$ is learned and padding positions are
excluded from the softmax.

\subsection{Macro-GT: Ecosystem Context Transformer}
\label{sec:method:macro}

Macro-GT models the global ecosystem context
$\mathbf{C}_{\tau(v_0)}$ available when target session $v_0$ is scored.
For each calendar date $t$, we construct typed climate tokens summarizing
complementary aspects of the fraud ecosystem:
(1) global fraud activity ($G$), including multi-window fraud-rate trends,
session volume, and acceleration;
(2) platform-specific behavior ($P_1,P_2$), capturing per-channel fraud rates,
traffic share, and temporal trends;
(3) cross-platform shifts ($X$), reflecting changes in fraud concentration
across channels;
(4) velocity and anomaly signals ($V$), capturing sudden changes in fraud rate,
volume, and campaign intensity;
(5) ecosystem topology ($E$), summarizing new-entity activity, identifier
sharing, and infrastructure concentration; and
(6) the current risk regime ($R$), including label coverage, high-risk
infrastructure prevalence, and repeat-fraud activity.
All statistics are computed non-anticipatively; in particular, label-dependent
quantities use only sessions satisfying $\tau(u)+\ell\leq t$.
Complete climate-feature definitions are provided in
Appendix~\ref{app:climate_tokens}.

Each climate feature vector is projected to $\mathbb{R}^{d}$ and combined
with a learned type embedding. Macro-GT additionally receives the target
session representation, its time-conditioned identifier states, and adaptive
class prototypes representing characteristic fraud and benign patterns.
Let $\mathbf{s}_0$ denote the projected target representation,
$\mathbf{c}_k$ a climate token, and $\mathbf{i}_r$ the encoded state
$i_{r,a_r}^{(\tau(v_0))}$ of the target's identifier of type $r$, where
$a_r=\operatorname{id}_r(v_0)$. The input sequence is
\begin{equation}
    \mathbf{S}^{(0)}
    =
    \left[
        \mathbf{s}_0
        \;\Big|\;
        \mathbf{c}_G,\mathbf{c}_{P_1},\mathbf{c}_{P_2},
        \mathbf{c}_X,\mathbf{c}_V,\mathbf{c}_E,\mathbf{c}_R
        \;\Big|\;
        \mathbf{i}_1,\ldots,\mathbf{i}_{|\mathcal{R}|}
        \;\Big|\;
        \boldsymbol{\mu}_1^+,\ldots,\boldsymbol{\mu}_{\Kf}^+,
        \boldsymbol{\mu}_1^-,\ldots,\boldsymbol{\mu}_{\Knf}^-
    \right].
    \label{eq:macro_sequence}
\end{equation}

A standard pre-norm transformer layer applies full self-attention over
$\mathbf{S}^{(0)}$, allowing the target to jointly compare its own state with
the current ecosystem context, its identifier history, and representative
fraud and benign patterns. The resulting macro representation
$\mathbf{h}_{\mathrm{macro}}\in\mathbb{R}^{d}$ is obtained using the same
seed-conditioned attention pooling as in Micro-GT, with $\mathbf{s}_0$ as the
query.

\subsubsection{Adaptive Class Prototypes}
\label{sec:method:centroids}

Fraud and benign sessions may exhibit multiple characteristic patterns rather
than forming a single cluster. We therefore maintain separate sets of
representative prototypes in the Macro-GT embedding space:
\begin{equation}
    \mathcal{P}^{+}
    =
    \left\{
        \boldsymbol{\mu}_k^{+}
    \right\}_{k=1}^{\Kf},
    \qquad
    \mathcal{P}^{-}
    =
    \left\{
        \boldsymbol{\mu}_k^{-}
    \right\}_{k=1}^{\Knf},
    \label{eq:prototype_sets}
\end{equation}
where $\mathcal{P}^{+}$ and $\mathcal{P}^{-}$ provide reference patterns for
fraud and benign sessions, respectively.

\textit{Initialization.}
The two prototype sets are initialized separately using K-Means++ seeding
\citep{arthur2007kmeans} on class-stratified training embeddings in
batch-normalized representation space.

\textit{Adaptive update.}
After each training batch, a session embedding is assigned only to the nearest
prototype of its own class. For class $c\in\{+,-\}$, let
$\mathcal{C}_k^{c}$ denote the batch embeddings assigned to prototype
$\boldsymbol{\mu}_k^{c}$. Assigned prototypes are updated by exponential
moving average (EMA):
\begin{equation}
    \boldsymbol{\mu}_k^{c}
    \leftarrow
    \alpha\,\boldsymbol{\mu}_k^{c}
    +
    (1-\alpha)
    \frac{1}{|\mathcal{C}_k^{c}|}
    \sum_{v\in\mathcal{C}_k^{c}}
    \hat{\mathbf{h}}_{\mathrm{macro}}(v),
    \qquad
    c\in\{+,-\},
    \label{eq:prototype_update}
\end{equation}
where
$\hat{\mathbf{h}}_{\mathrm{macro}}
=\operatorname{BN}(\mathbf{h}_{\mathrm{macro}})$ and
$\alpha\in(0,1)$ is the EMA momentum. The prototypes are maintained as
non-learnable buffers and updated outside the backward pass, decoupling their
adaptation from the classification objective. Prototypes that remain
unassigned beyond a fixed patience threshold are reinitialized from
class-matched batch embeddings to maintain coverage of the representation
space.


\subsection{Training Objective}
\label{sec:method:training}

The micro and macro embeddings are combined via a learned linear projection:
\begin{equation}
    \hout = \mathbf{W}_c \left[\hmicro \;\|\; \hmacro\right] + \mathbf{b}_c,
\end{equation}
and a two-layer MLP head with GELU activation and dropout produces the scalar logit.
\model{} is trained end-to-end with class-reweighted binary cross-entropy:
\begin{equation}
    \mathcal{L} = -\frac{1}{|\mathcal{B}|} \sum_{v \in \mathcal{B}}
    \left[w^+\, y_v \log \hat{y}_v + (1-y_v) \log (1 - \hat{y}_v)\right],
    \quad w^+ = \frac{|\mathcal{V}_0|}{|\mathcal{V}_1|},
\end{equation}
where $\hat{y}_v = \sigma(\mathrm{logit}_v)$, $|\mathcal{V}_0|$ and $|\mathcal{V}_1|$
are the non-fraud and fraud counts in the training set, and $\mathcal{B}$ is a mini-batch
sampled via a neighborhood loader.
The EMA prototype update is applied after each gradient step, strictly outside the backward
pass.

\section{Experiments}
\label{sec:experiments}

\subsection{Dataset and Evaluation Protocol} \label{sec:exp:setup} We evaluate \model{} on a proprietary dataset of more than 130 million high-risk transaction sessions collected over one year from the digital channels of a leading U.S. financial institution. Each session includes behavioral, platform, and device-related features, together with shared account, device, and network identifiers that define the relational graph. Fraud labels are binary and become observable only after an adjudication delay of $\ell$ days. Fraud accounts for less than 0.1\% of all sessions, resulting in extreme class imbalance. We use a chronological split: the final four months form the test set, the preceding two months form the validation set, and all earlier sessions are used for training. The four test months are also evaluated separately for temporal-stability analysis. All models use the same temporal splits and obey the serve-time and label-maturity constraints defined in Section~\ref{sec:problem}, preventing future-session and immature-label leakage. We report ROC-AUC overall and separately for two digital access channels, denoted $P_1$ and $P_2$ for confidentiality. To evaluate deployment-relevant performance, we additionally report two operating-point metrics based on the \emph{flag rate}, i.e., the fraction of sessions flagged as high risk and subjected to friction interventions. Flag rate therefore serves as an operational proxy for customer friction. Specifically, \emph{FR@50R} denotes the flag rate required to achieve 50\% fraud recall (lower is better), while \emph{R@5FR} denotes fraud recall at a 5\% flag rate (higher is better).

\subsection{Results} 
\label{sec:exp:results} 
Table~\ref{tab:ablation2} compares \model{} with production, tabular, sequential, and graph-based baselines, followed by an incremental ablation of the proposed architecture. All relative changes are reported with respect to the production XGBoost system.

\paragraph{Baseline comparison.}
\textit{XGBoost (production)} is the gradient-boosted production system that scores sessions independently and serves as the primary reference, achieving 80.26\% overall AUC, 78.05\% P1 AUC, and 74.10\% P2 AUC. \textit{MLP} performs nearly identically, indicating that additional model capacity alone provides little benefit without relational information. \textit{FT-Transformer}~\citep{gorishniy2021revisiting} underperforms XGBoost, particularly on P2 (66.54\% AUC). \textit{Transformer-Seq} improves overall AUC to 81.95\% but remains weak on P2 (72.97\%), showing that sequential context alone cannot capture broader cross-session dependencies. \textit{GNN (GraphSAGE)} exploits the shared-identifier graph, increasing overall AUC to 82.00\% and P2 AUC to 76.20\%, while reducing FR@50R by 10.03\% relative to production. \textit{GNN + LP (ATLAS)}~\citep{kerdabadi2025spatio} further incorporates lag-safe fraud history from connected predecessors, reaching 84.02\% overall AUC, 82.45\% P1 AUC, and 78.75\% P2 AUC. Relative to production, ATLAS reduces FR@50R by 34.59\% and improves R@5FR by 19.62\%, establishing a strong relational baseline and highlighting the value of adjudicated fraud signals from connected sessions.

\paragraph{Component ablation.}
Starting from ATLAS, \textit{Micro-GT (full attention)} replaces homogeneous SAGE-style aggregation with a local transformer over the heterogeneous node tokens defined in Eq.~\eqref{eq:token_components}. This raises overall AUC from 84.02\% to 84.58\% and P2 AUC from 78.75\% to 80.47\%, while improving FR@50R from 6.325\% to 5.821\% and R@5FR from 46.93\% to 47.75\%. These gains show that incorporating node heterogeneity and direct all-pair interactions captures higher-order dependencies beyond conventional message passing. \textit{Micro-GT (relational masks)} further increases overall AUC to 84.78\% and P1 AUC to 83.10\%, while maintaining comparable operational performance, supporting structured attention over account, device, network, and multi-hop relational interactions. Adding \textit{Macro-GT (climate tokens)} raises overall AUC to 84.93\% and P2 AUC to 81.06\%, while reducing FR@50R to 5.519\% and increasing R@5FR to 48.44\%. The strong P2 improvement indicates that ecosystem-level fraud dynamics, platform shifts, velocity, and infrastructure signals provide complementary information beyond the local neighborhood. Finally, adaptive class prototypes complete \model{}, yielding the best performance across all reported metrics: 85.12\% overall AUC, 83.46\% P1 AUC, 81.35\% P2 AUC, 5.373\% FR@50R, and 48.91\% R@5FR. Compared with production, \model{} improves overall, P1, and P2 AUC by 4.86, 5.41, and 7.25 points, respectively, while achieving a 44.44\% relative reduction in FR@50R and a 24.66\% relative improvement in R@5FR. Overall, the results demonstrate complementary gains from expressive local relational modeling, structured relation-aware attention, global ecosystem context, and adaptive class prototypes.

\begin{table}[t]
\centering
\footnotesize
\setlength{\tabcolsep}{2.5pt}
\renewcommand{\arraystretch}{1.05}

\caption{Overall performance and component ablation. All values are in \%.
FR@50R denotes the flag rate at 50\% fraud recall ($\downarrow$), and R@5FR
denotes fraud recall at a 5\% flag rate ($\uparrow$). Parentheses report
relative change from the production XGBoost baseline.}
\label{tab:ablation2}
\smallskip

\begin{tabular}{@{}lrrrrr@{}}
\toprule
\textbf{Model}
& \textbf{AUC}
& \textbf{P1 AUC}
& \textbf{P2 AUC}
& \textbf{FR@50R$\downarrow$}
& \textbf{R@5FR$\uparrow$} \\
\midrule

XGBoost (production)
& 80.26
& 78.05
& 74.10
& 9.671
& 39.24 \\

MLP
& 80.29 \tiny{(+0.03)}
& 77.99 \tiny{($-$0.08)}
& 75.10 \tiny{(+1.35)}
& 9.360 \tiny{($-$3.21)}
& 39.78 \tiny{(+1.39)} \\

FT-Transformer
& 79.16 \tiny{($-$1.37)}
& 77.02 \tiny{($-$1.31)}
& 66.54 \tiny{($-$10.20)}
& 11.213 \tiny{(+15.94)}
& 35.42 \tiny{($-$9.74)} \\

Transformer-Seq
& 81.95 \tiny{(+2.10)}
& 80.28 \tiny{(+2.86)}
& 72.97 \tiny{($-$1.52)}
& 9.244 \tiny{($-$4.41)}
& 39.60 \tiny{(+0.91)} \\

GraphSAGE
& 82.00 \tiny{(+2.16)}
& 80.05 \tiny{(+2.57)}
& 76.20 \tiny{(+2.84)}
& 8.701 \tiny{($-$10.03)}
& 41.54 \tiny{(+5.87)} \\

GraphSAGE + LP (ATLAS)
& 84.02 \tiny{(+4.69)}
& 82.45 \tiny{(+5.64)}
& 78.75 \tiny{(+6.28)}
& 6.325 \tiny{($-$34.59)}
& 46.93 \tiny{(+19.62)} \\

\addlinespace[0.5mm]
\cmidrule(lr){1-6}
\addlinespace[0.3mm]

Micro-GT (full attention)
& 84.58 \tiny{(+5.39)}
& 82.83 \tiny{(+6.13)}
& 80.47 \tiny{(+8.59)}
& 5.821 \tiny{($-$39.80)}
& 47.75 \tiny{(+21.70)} \\

Micro-GT (relational masks)
& 84.78 \tiny{(+5.64)}
& 83.10 \tiny{(+6.48)}
& 80.37 \tiny{(+8.47)}
& 5.801 \tiny{($-$40.01)}
& 47.83 \tiny{(+21.89)} \\

+ Macro-GT
& 84.93 \tiny{(+5.82)}
& 83.24 \tiny{(+6.65)}
& 81.06 \tiny{(+9.40)}
& 5.519 \tiny{($-$42.93)}
& 48.44 \tiny{(+23.45)} \\

\rowcolor{gray!10}
\textbf{\model{} (+centroids)}
& \textbf{85.12} \tiny{(+6.06)}
& \textbf{83.46} \tiny{(+6.94)}
& \textbf{81.35} \tiny{(+9.79)}
& \textbf{5.373} \tiny{($-$44.44)}
& \textbf{48.91} \tiny{(+24.66)} \\

\bottomrule
\end{tabular}
\vspace{-4mm}
\end{table}

\subsection{Temporal Stability}
\label{sec:exp:temporal}

\begin{figure}[t]
    \centering
    \includegraphics[width=0.70\linewidth]{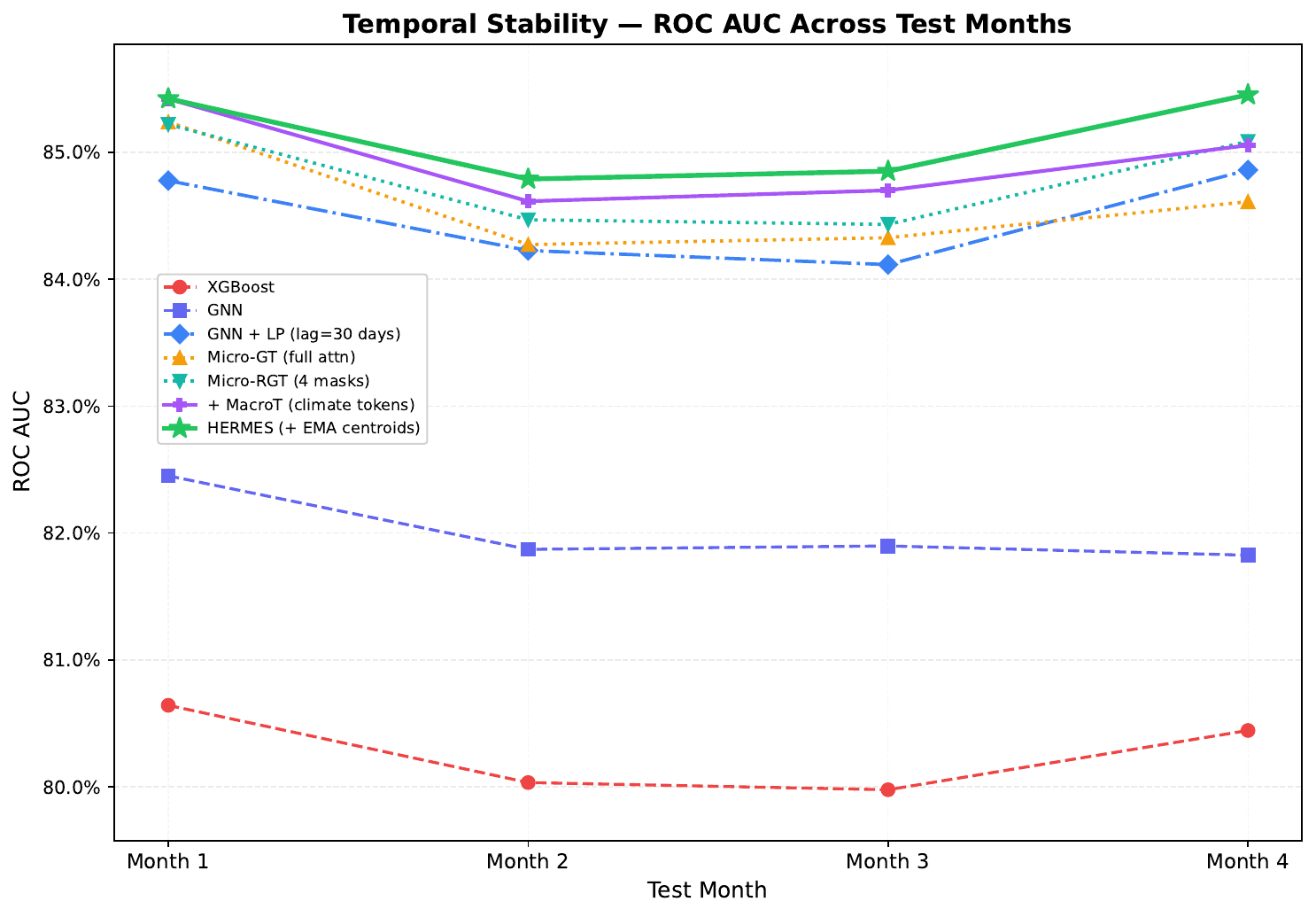}
    \caption{Temporal stability: ROC-AUC of \model{} and key baselines across four
consecutive held-out test months. \model{} consistently outperforms all baselines.}
    \label{fig:temporal}
\end{figure}

To assess robustness to temporal distribution shift, we evaluate \model{} and key
baselines separately across four consecutive held-out test months.
Figure~\ref{fig:temporal} shows that \model{} consistently outperforms all
baselines throughout the evaluation window, with no degradation relative to the
graph-based models in later months. The gains remain stable across periods with
changing fraud activity, indicating that the combination of local relational
modeling and global ecosystem context improves robustness across evolving fraud
regimes. Such temporal stability is particularly important in production settings,
where fraud distributions shift continuously, and frequent model recalibration is costly.

\section{Related Work}
\label{sec:related}
\vspace{-2mm}

Account takeover detection in operational systems requires balancing fraud prevention against unnecessary intervention on legitimate users. \citet{doerfler2019evaluating} study this tradeoff at Google by evaluating risk-aware login challenges against real-world account hijacking attempts and quantifying the resulting user friction. In an industrial marketplace setting, \citet{kawase2019internet} analyze ATO at mobile.de and develop a concept-drift-sensitive detection strategy for evolving attacker behavior.

Relational modeling has also been explored more broadly for financial fraud \citep{liu2018heterogeneous, jiang2022spade, egressy2024provably}. TitAnt~\citep{10.14778/3352063.3352126} demonstrates large-scale real-time transaction fraud detection using transaction-network information at Ant Financial, while xFraud~\citep{10.14778/3494124.3494128} applies heterogeneous graph neural networks to large-scale real-world transaction networks. GTAN~\citep{xiang2023semi} constructs temporal transaction graphs and propagates risk through gated temporal attention, and FraudGT~\citep{lin2024fraudgt} introduces an edge-aware graph transformer for financial transaction graphs. 
Beyond fraud-specific architectures, Relational Graph Transformer
(RelGT)~\citep{ICLR2026_fa1ef433} introduces multi-element relational
tokenization with local subgraph attention and global centroids for general
relational prediction. \model{} builds on this tokenization framework while
introducing temporally structured local attention, ecosystem climate tokens,
and class-stratified EMA prototypes for non-anticipative ATO detection.
Among ATO-specific methods, ATLAS~\citep{kerdabadi2025spatio} is most closely
related, modeling consumer-banking ATO as node classification on a
time-respecting session graph that links sessions through shared account,
device, and network identifiers using GraphSAGE and serve-time-safe label
propagation. Building on this line of relational fraud modeling, \model{}
combines structured local reasoning over heterogeneous, time-conditioned
session relationships with non-anticipative ecosystem-level context and
adaptive fraud and benign reference prototypes.



\section{Conclusion}
\label{sec:conclusion}
\vspace{-2mm}

We presented \model{}, a dual-scale graph transformer architecture for ATO fraud detection that combines structured local relational modeling with global ecosystem context. \emph{Micro-GT} captures heterogeneous multi-hop interactions, while \emph{Macro-GT} incorporates non-anticipative climate signals and adaptive class prototypes. Evaluated on more than 130 million high-risk transaction sessions, \model{} achieves 85.12\% ROC-AUC, a 44.44\% relative reduction in customer friction, and a 24.66\% relative improvement in fraud recall over the production system. Ablation and temporal analyses show complementary gains from the local and global components and robust performance across evolving fraud conditions.

\bibliographystyle{plainnat}
\bibliography{references}


\appendix

\section{Appendix}

\subsection{Token Feature Definitions}
\label{app:tokens}

Throughout this appendix, $\epsilon > 0$ denotes a small constant for numerical stability,
$\mathrm{clip}(x, a, b) = \min(\max(x, a), b)$, $y_u \in \{0,1\}$ is the fraud label of session $u$,
$\tau(u)$ its timestamp, $\ell$ the label lag, $t_0$ the dataset start date,
and $\operatorname{id}_r(u)$ the identifier of type $r \in \mathcal{R}$ associated with session $u$.
We use $\operatorname{acc}(u)$, $\operatorname{dev}(u)$, and $\operatorname{net}(u)$ as shorthand
for the account, device, and network identifiers, respectively.

\subsection{Label-Propagated (LP) Features}
\label{app:lp_features}

For each session $v$, let
\begin{equation}
    \mathcal{L}^{-}(v)
    =
    \bigcup_{r \in \mathcal{R}} \mathcal{L}_{r}^{-}(v)
\end{equation}
denote the set of direct predecessors with adjudicated labels available at score time,
where $\mathcal{L}_{r}^{-}(v)$ is defined in Section~\ref{sec:problem}.
Four aggregate statistics are computed and prepended to $\mathbf{x}_v$:
\begin{align}
    n_{\ell}(v) &= |\mathcal{L}^{-}(v)| && \text{number of labeled predecessors,} \\
    n_f(v) &= \sum_{u \in \mathcal{L}^{-}(v)} y_u && \text{count of fraudulent predecessors,} \\
    \rho_f(v) &= \frac{n_f(v)}{\max(n_{\ell}(v),\, 1)} && \text{local fraud rate,} \\
    \mathbb{1}_f(v) &= \mathbf{1}[n_f(v) > 0] && \text{any-fraud indicator.}
\end{align}

\subsection{Identifier Token Features}
\label{app:id_tokens}

For relation type $r$ and identifier $a$, each time-conditioned identifier state
$i_{r,a}^{(t)}$ carries a 25-dimensional feature vector representing the information
available for identifier $a$ at time $t$. Let
\begin{equation}
    \mathcal{A}_{r,a}(t)
    =
    \{u \in \mathcal{S} :
    \operatorname{id}_r(u)=a,\;
    \tau(u)<t\}
\end{equation}
denote all sessions observed through identifier $a$ before $t$, and let
\begin{equation}
    \mathcal{A}_{r,a}^{\ell}(t)
    =
    \{u \in \mathcal{A}_{r,a}(t) :
    \tau(u)+\ell\leq t\}
\end{equation}
denote the lag-safe subset. For readability, we suppress the identifier index $a$
from the feature symbols below.

\subsection{Fraud rates (lag-safe, 4 features):}
\begin{equation}
    \rho_r^{(w)}(t) =
    \frac{
        \sum_{u \in \mathcal{A}_{r,a}^{\ell}(t),\; \tau(u) \geq t-w} y_u
    }{
        \max\bigl(
        |\{u \in \mathcal{A}_{r,a}^{\ell}(t) : \tau(u)\geq t-w\}|,\;1
        \bigr)
    },
    \quad
    w \in \{7,30,90,\infty\}\text{ days}.
\end{equation}

\paragraph{Absolute fraud and clean counts (lag-safe, 2 features):}
\begin{align}
    n_f^r(t)
    &= \sum_{u \in \mathcal{A}_{r,a}^{\ell}(t)} y_u,
    \qquad
    n_c^r(t)
    = |\mathcal{A}_{r,a}^{\ell}(t)| - n_f^r(t).
\end{align}

\paragraph{Session volume (no lag, 4 features):}
\begin{equation}
    c_r^{(w)}(t)
    =
    |\{u \in \mathcal{A}_{r,a}(t) : \tau(u)\geq t-w\}|,
    \quad
    w \in \{7,30,90,\infty\}\text{ days}.
\end{equation}

\paragraph{Maturity (no lag, 2 features):}
\begin{align}
    \text{days\_active}_r(t)
    &= t - \min_{u \in \mathcal{A}_{r,a}(t)} \tau(u), \\
    \text{recency}_r(t)
    &= t - \max_{u \in \mathcal{A}_{r,a}(t)} \tau(u).
\end{align}

\paragraph{Co-identifier diversity (no lag, 6 features):}
Let $r'$ and $r''$ be the two identifier types other than $r$.
For window $w \in \{7,30,90\}$ days, count distinct co-occurring identifiers
of type $r'$ observed alongside $a$:
\begin{equation}
    d_{r,r'}^{(w)}(t)
    =
    \bigl|
    \bigl\{
        \operatorname{id}_{r'}(u) :
        u \in \mathcal{A}_{r,a}(t),\;
        \tau(u)\geq t-w
    \bigr\}
    \bigr|.
\end{equation}
Features are computed for both $r'$ and $r''$, giving $2\times3=6$ features.

\paragraph{Fraud-linked co-identifier counts (lag-safe, 4 features):}
For window $w \in \{30,90\}$ days and each of the two other identifier types
$r',r''$:
\begin{equation}
    d_{r,r'}^{f,(w)}(t)
    =
    \bigl|
    \bigl\{
        \operatorname{id}_{r'}(u) :
        u \in \mathcal{A}_{r,a}^{\ell}(t),\;
        y_u=1,\;
        \tau(u)\geq t-w
    \bigr\}
    \bigr|.
\end{equation}

\paragraph{Coverage (3 features):}
\begin{equation}
    \text{cov}_r^{(w)}(t)
    =
    \frac{\min(\text{days\_active}_r(t),\,w)}{w},
    \quad
    w \in \{7,30,90\}\text{ days}.
\end{equation}

\subsection{Cross-Identifier Session Features}
\label{app:cross_tokens}

Each session node $v$ is further augmented with 7 cross-identifier features appended to
$\mathbf{x}_v$, capturing pairwise relational risk signals across identifier types.
Let
\begin{equation}
    \mathcal{P}_{r,r'}(v,t)
    =
    \left\{
        u \in \mathcal{S} :
        \operatorname{id}_r(u)=\operatorname{id}_r(v),\;
        \operatorname{id}_{r'}(u)=\operatorname{id}_{r'}(v),\;
        \tau(u)<t
    \right\}
\end{equation}
denote sessions sharing both identifiers $r$ and $r'$ with $v$ before $t$, and let
\begin{equation}
    \mathcal{P}_{r,r'}^{\ell}(v,t)
    =
    \{u \in \mathcal{P}_{r,r'}(v,t) :
    \tau(u)+\ell\leq t\}
\end{equation}
denote the lag-safe subset. $W$ denotes the lookback window for pair-level features.
For quantities inherited from the identifier features above, the identifier for relation
$r$ is $a_r=\operatorname{id}_r(v)$.

\begin{align}
    \rho_{r,r'}(v,t)
    &=
    \frac{
        \sum_{u \in \mathcal{P}_{r,r'}^{\ell}(v,t)} y_u
    }{
        \max(|\mathcal{P}_{r,r'}^{\ell}(v,t)|,1)
    },
    \quad
    (r,r') \in
    \{(\text{acc},\text{dev}),(\text{acc},\text{net}),(\text{dev},\text{net})\},
    \\
    n_f^{\text{pair}}(v,t)
    &=
    \sum_{u \in \mathcal{P}_{\text{acc},\text{dev}}^{\ell}(v,t)} y_u,
    \\
    \rho_{\max}(v,t)
    &=
    \max_{r \in \mathcal{R}} \rho_r^{(90)}(t),
    \\
    \mathbb{1}_{\text{any}}(v,t)
    &=
    \mathbf{1}
    \bigl[
        \exists\,r \in \mathcal{R} :
        n_f^r(t)>0
    \bigr],
    \\
    \text{cov}_{\text{pair}}(v,t)
    &=
    \frac{\min(t-t_0,\,W)}{W}.
\end{align}

\subsection{Ecosystem Climate Token Features}
\label{app:climate_tokens}

Climate tokens are precomputed daily at date $t$.
Let
$\mathcal{D}(t,w)=\{u \in \mathcal{S}:t-w\leq\tau(u)<t\}$
denote all sessions in a $w$-day window before $t$, and let
$\mathcal{D}^{\ell}(t,w)=
\{u \in \mathcal{D}(t,w):\tau(u)+\ell\leq t\}$
denote the lag-safe subset.
The dataset start date $t_0$ is used for coverage normalization.
For platform $p \in \{P_1,P_2\}$, let
$\mathcal{D}_p(t,w)=
\{u \in \mathcal{D}(t,w):\operatorname{platform}(u)=p\}$
and let $\mathcal{D}_p^{\ell}(t,w)$ denote the corresponding lag-safe subset.
Define the global fraud rate
\[
\rho(t,w)
=
\frac{\sum_{u \in \mathcal{D}^{\ell}(t,w)}y_u}
{\max(|\mathcal{D}^{\ell}(t,w)|,1)}
\]
and the platform fraud rate
\[
\rho_p(t,w)
=
\frac{\sum_{u \in \mathcal{D}_p^{\ell}(t,w)}y_u}
{\max(|\mathcal{D}_p^{\ell}(t,w)|,1)}.
\]

\paragraph{Token G --- Global fraud climate (12 features):}
\begin{align}
    \rho_w^G
    &= \rho(t,w),
    \quad
    w \in \{7,14,30,90\}\text{ days},
    \\
    \nu_w^G
    &= \log(1+|\mathcal{D}(t,w)|),
    \quad
    w \in \{7,30\}\text{ days},
    \\
    \delta_{7/30}^G
    &= \rho(t,7)/(\rho(t,30)+\epsilon)-1,
    \quad
    \delta_{30/90}^G
    = \rho(t,30)/(\rho(t,90)+\epsilon)-1,
    \\
    a^G
    &= \delta_{7/30}^G-\delta_{30/90}^G
    \quad \text{(fraud acceleration)},
    \\
    \text{cov}_w^G
    &= \min(t-t_0,w)/w,
    \quad
    w \in \{7,30,90\}.
\end{align}

\paragraph{Tokens $P_1$, $P_2$ --- Platform-specific climate (10 features each):}
For platform $p$:
\begin{align}
    \rho_w^p
    &= \rho_p(t,w),
    \quad
    w \in \{7,30,90\},
    \\
    \sigma_7^p
    &= |\mathcal{D}_p(t,7)|/
    \max(|\mathcal{D}(t,7)|,1)
    \quad \text{(session share)},
    \\
    \phi_7^p
    &=
    \sum_{u \in \mathcal{D}_p^{\ell}(t,7)}y_u
    \;/\;
    \max\!\left(
        \sum_{u \in \mathcal{D}^{\ell}(t,7)}y_u,
        1
    \right)
    \quad \text{(fraud share)},
    \\
    \delta_{7/30}^p
    &= \rho_p(t,7)/(\rho_p(t,30)+\epsilon)-1,
    \quad
    \nu_7^p=\log(1+|\mathcal{D}_p(t,7)|),
    \\
    \text{cov}_w^p
    &= \min(t-t_0,w)/w,
    \quad
    w \in \{7,30,90\}.
\end{align}

\paragraph{Token X --- Cross-platform shift (5 features):}
Let
\[
\phi_w^p
=
\frac{
    \sum_{u \in \mathcal{D}_p^{\ell}(t,w)}y_u
}{
    \max(
        \sum_{u \in \mathcal{D}^{\ell}(t,w)}y_u,
        1
    )
}
\]
denote the share of total fraud attributable to platform $p$ in window $w$, and let
\[
\sigma_w^p
=
|\mathcal{D}_p(t,w)|/
\max(|\mathcal{D}(t,w)|,1)
\]
denote its session share.
\begin{align}
    \xi_w
    &=
    \rho_{P_1}(t,w)/
    (\rho_{P_2}(t,w)+\epsilon),
    \quad
    w \in \{7,30\},
    \\
    \gamma_w
    &=
    \phi_w^{P_1}-\sigma_w^{P_1},
    \quad
    w \in \{7,30\}
    \quad \text{(fraud concentration shift)},
    \\
    H_7
    &=
    -\sigma_7^{P_1}\log(\sigma_7^{P_1}+\epsilon)
    -(1-\sigma_7^{P_1})
    \log(1-\sigma_7^{P_1}+\epsilon).
\end{align}

\paragraph{Token V --- Velocity and anomaly (7 features):}
\begin{align}
    z_\rho^{(w)}
    &=
    \mathrm{clip}\!\left(
        \frac{
            \rho(t,w)-\rho(t,90)
        }{
            \sqrt{
                \rho(t,90)(1-\rho(t,90))+\epsilon
            }
        },
        -5,5
    \right),
    \quad
    w \in \{7,30\},
    \\
    z_\nu^{(7)}
    &=
    \mathrm{clip}\!\left(
        \frac{
            |\mathcal{D}(t,7)|/7
            -
            |\mathcal{D}(t,90)|/90
        }{
            \sqrt{|\mathcal{D}(t,90)|/90+\epsilon}
        },
        -5,5
    \right),
    \\
    z_f^{(7)}
    &=
    \mathrm{clip}\!\left(
        \frac{
            f_7/7-f_{90}/90
        }{
            \sqrt{f_{90}/90+\epsilon}
        },
        -5,5
    \right),
    \quad
    f_w
    =
    \sum_{u \in \mathcal{D}^{\ell}(t,w)}y_u,
    \\
    b^G
    &=
    \mathrm{clip}\!\left(
        \frac{
            \max_{k=0}^{6}\rho(t-k,1)
        }{
            \rho(t,30)+\epsilon
        },
        0,20
    \right)
    \quad \text{(fraud burst)},
    \\
    \text{cov}_w^V
    &=
    \min(t-t_0,w)/w,
    \quad
    w \in \{7,30\}.
\end{align}

\paragraph{Token E --- Ecosystem topology (10 features):}
Let $\mathcal{X}_r(t,w)$ be the set of unique identifiers of type $r$ observed in
$\mathcal{D}(t,w)$, and let
$\mathcal{X}_r^{\text{new}}(t,w)\subseteq\mathcal{X}_r(t,w)$ denote those first
observed within $[t-w,t)$.
\begin{align}
    \eta_r^{(w)}
    &=
    |\mathcal{X}_r^{\text{new}}(t,w)|
    /
    \max(|\mathcal{X}_r(t,w)|,1),
    \quad
    r \in \{\text{dev},\text{acc}\},
    \; w=7,
    \\
    \bar{s}_{\text{dev}}(t)
    &=
    \text{mean}_{i \in \mathcal{X}_{\text{dev}}(t,7)}
    \left|
    \left\{
        j \in \mathcal{X}_{\text{acc}}(t,7) :
        \exists u,\;
        \operatorname{dev}(u)=i
        \wedge
        \operatorname{acc}(u)=j
    \right\}
    \right|,
    \\
    \kappa_{\text{net}}(t)
    &=
    \sum_{\text{top-5}} c_i
    \;/\;
    |\mathcal{D}(t,7)|,
    \quad
    c_i
    =
    |\{u \in \mathcal{D}(t,7):
    \operatorname{net}(u)=i\}|,
    \\
    \psi(t)
    &=
    |\mathcal{X}_{\text{acc}}(t,7)|
    /
    \max(|\mathcal{X}_{\text{dev}}(t,7)|,1),
    \\
    \Delta\eta_{\text{dev}}
    &=
    \mathrm{clip}\!\left(
        \eta_{\text{dev}}^{(7)}
        /
        (\eta_{\text{dev}}^{(30)}+\epsilon)-1,
        -2,5
    \right),
    \\
    \Delta\kappa_{\text{net}}
    &=
    \mathrm{clip}\!\left(
        |\mathcal{X}_{\text{net}}(t,7)|
        /
        (7|\mathcal{X}_{\text{net}}(t,30)|/30+\epsilon)-1,
        -2,5
    \right),
    \\
    \text{cov}_w^E
    &=
    \min(t-t_0,w)/w,
    \quad
    w \in \{7,30,90\}.
\end{align}

\paragraph{Token R --- Risk regime (7 features):}
Let
\[
\mathcal{F}^{\ell}(t)
=
\{u \in \mathcal{S} :
\tau(u)+\ell\leq t,\;
y_u=1\}
\]
be all confirmed fraud sessions before $t$, and define
\[
h_r^{(w)}(t)
=
\frac{
\left|
\left\{
u \in \mathcal{D}(t,w) :
\operatorname{id}_r(u)
\in
\{\operatorname{id}_r(f):f\in\mathcal{F}^{\ell}(t)\}
\right\}
\right|
}{
\max(|\mathcal{D}(t,w)|,1)
}
\]
as the fraction of sessions in window $w$ that share an identifier of type $r$
with any known-fraud session.
\begin{align}
    \lambda(t)
    &=
    |\mathcal{D}^{\ell}(t,30)|
    /
    \max(|\mathcal{D}(t,30)|,1)
    \quad \text{(label coverage)},
    \\
    h_r^{(7)}(t)
    &\quad
    r \in \{\text{net},\text{dev}\}
    \quad \text{(high-risk identifier share, 7-day)},
    \\
    \chi_{\text{acc}}(t)
    &=
    h_{\text{acc}}^{(7)}(t)
    \quad \text{(repeat-fraud account rate)},
    \\
    \Delta h_{\text{dev}}
    &=
    \mathrm{clip}\!\left(
        h_{\text{dev}}^{(7)}(t)
        /
        (h_{\text{dev}}^{(30)}(t)+\epsilon)-1,
        -2,5
    \right)
    \quad \text{(contamination trend)},
    \\
    \text{cov}_w^R
    &=
    \min(t-t_0,w)/w,
    \quad
    w \in \{7,30\}.
\end{align}

\end{document}